%% file: main.tex
\documentclass[conference,a4paper]{IEEEtran}
\IEEEoverridecommandlockouts

\usepackage[hidelinks]{hyperref}
\usepackage[all]{hypcap}
\usepackage[cmex10]{amsmath}
\usepackage{amssymb,amsfonts}
\usepackage{dblfloatfix}
\usepackage{graphicx}
\graphicspath{{figures/}}
\usepackage{booktabs}
\usepackage[numbers,compress]{natbib}
\usepackage{bm}
\usepackage{orcidlink}

\hypersetup{
    pdftitle={Remote-Sensing City Layout Extraction with MLLM},
    pdfauthor={Zigan Zhou, Kai Li, Yupeng Deng},
    pdfsubject={Executable city-layout extraction from remote-sensing imagery},
    pdfkeywords={remote sensing, multimodal large language models, city layout, code generation},
    bookmarksnumbered=true
}

\newcommand{\ZiganORCID}{\orcidlink{0009-0004-5184-709X}}
\newcommand{\KaiORCID}{\orcidlink{0000-0001-7161-434X}}
\newcommand{\YupengORCID}{\orcidlink{0000-0001-8396-0338}}

\begin{document}

\title{\uppercase{Remote-Sensing City Layout Extraction with MLLM}}

\author{
    \IEEEauthorblockN{Zigan Zhou\,\ZiganORCID}
    \IEEEauthorblockA{\textit{City University of Hong Kong}\\
    Hong Kong SAR, China\\
    ziganzhou4-c@my.cityu.edu.hk}
    \and
    \IEEEauthorblockN{Kai Li\,\KaiORCID}
    \IEEEauthorblockA{\textit{City University of Hong Kong}\\
    Hong Kong SAR, China\\
    \textit{School of Electronic, Electrical and}\\
    \textit{Communication Engineering}\\
    \textit{University of Chinese Academy of Sciences}\\
    Beijing, China\\
    likai211@mails.ucas.ac.cn}
    \and
    \IEEEauthorblockN{Yupeng Deng\,\YupengORCID}
    \IEEEauthorblockA{\textit{Aerospace Information Research Institute}\\
    \textit{Chinese Academy of Sciences}\\
    Beijing, China\\
    dengyp@aircas.ac.cn}
}

\maketitle

\input{chapters/00_abstract}
\input{chapters/01_introduction}
\input{chapters/02_related_work}
\input{chapters/03_method}
\input{chapters/04_experiments}
\input{chapters/05_conclusion}

\small
\bibliographystyle{IEEEtranN}
\bibliography{references}

\end{document}

%% file: chapters/00_abstract.tex
\begin{abstract}
Remote-sensing systems usually describe urban content with detection boxes, semantic masks, or vector boundaries. Such outputs locate classes and support image-plane scoring, yet they do not by themselves constitute an executable layout that retains object identities, typed relations, topology, and regeneration rules. The framework instead casts urban-layout extraction from a single top-down image as constrained code generation with a multimodal large language model (MLLM). An image model first produces an aligned five-class semantic layout prior. Three ordered MLLM passes use the image and this prior to recover roads, land-cover regions and relations, and buildings. Deterministic normalization converts the accumulated records into a city graph and a restricted layout program. Executing the program creates a renderable 3D city layout and an orthographic semantic projection over shared geometry. The projection admits pixel-level comparison with remote-sensing masks, while named objects, relations, and editing operations remain available for synchronized regeneration of both views. Evaluated on the 100 scenes of CityLayout-100, the complete framework obtains 41.1\% mean intersection-over-union and 48.3\% global intersection-over-union. This result provides quantitative evidence that visual observations can be translated into inspectable, editable city code with coupled planar and 3D outputs.
\end{abstract}


%% file: chapters/01_introduction.tex
\section{Introduction}

Detection and segmentation turn remote-sensing imagery into boxes, masks, or vector boundaries that specify what is present and where it is located. These are effective image-plane interfaces, but the usual outputs do not jointly preserve persistent identity, typed relations, topology, and rules for regeneration. That distinction becomes consequential when image interpretation is expected to feed a maintainable city asset. Structured formats and procedural models require explicit objects and construction logic, neither of which is directly available from one overhead view. This work asks whether a multimodal large language model (MLLM) can organize that visual evidence as an executable city layout without giving up segmentation-style measurement.

\begin{figure*}[t]
    \centering
    \includegraphics[width=\textwidth]{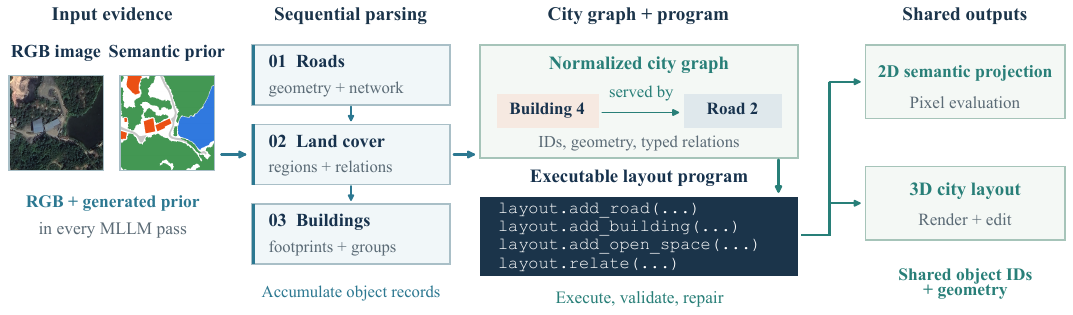}
    \caption{Overview of the proposed framework. A generated semantic layout prior guides the three MLLM parsing passes. The normalized city graph and executable program produce planar and 3D outputs with shared object identities and geometry.}
    \label{fig:pipeline}
\end{figure*}

The framework maps a near-nadir RGB image to a normalized city graph and a restricted layout program. A controller schedules three dependent passes---roads, land-cover regions and relations, and buildings---and transfers validated records between them. It subsequently executes the generated program and invokes fixed checks or a predefined repair routine when an artifact fails. Every MLLM pass receives a generated semantic layout prior with aligned class and footprint evidence, together with the RGB image for cues about identity, function, relations, appearance, and approximate height. The same named objects and geometry produce an orthographic semantic projection and a renderable 3D layout (Fig.~\ref{fig:pipeline}); the former is scored in the image plane, while the graph and program remain available for inspection, editing, and joint regeneration.

The work contributes an image-to-city-code task definition in which the primary prediction is an executable layout program rather than a terminal raster or vector. Named objects and typed relations survive execution, allowing planar and 3D outputs to be regenerated from one representation. It also introduces a prior-grounded MLLM workflow: decomposed spatial parsing is guided by a generated semantic layout prior, and validated intermediate records, constrained code, and predefined recovery paths leave a traceable generation record. For evaluation, CityLayout-100 couples standard pixel metrics on the orthographic projection with synchronized executable 3D examples over 100 scenes. The complete framework records 41.1\% mean IoU and 48.3\% global IoU, while Section~\ref{sec:prior-ablation} isolates the contribution of the semantic layout prior to spatial grounding.

%% file: chapters/02_related_work.tex
\section{Related Work}

Overhead scenes have been represented at several levels of geometric detail. Detection localizes oriented objects, instance segmentation separates individual masks, and semantic segmentation assigns dense classes \citep{xia2018dota,he2017maskrcnn,zamir2019isaid,chen2018deeplab}. IRSAMap carries this progression into vector land-cover objects and topology-aware mapping \citep{meng2025irsamap}. These task definitions serve localization and pixel evaluation, but they do not require a regenerable construction with persistent identities and typed relations. The framework preserves image-plane measurement through projection while placing executable structure in the prediction space.

Procedural and structured city representations approach the problem from the asset side. Context-sensitive shape grammars encode hierarchical construction rules for editable architecture \citep{muller2006procedural}, while CityJSON stores semantic objects, identifiers, attributes, and geometry in a machine-processable form \citep{ledoux2019cityjson}. In both cases, structure becomes useful once the rules or geometry are available. The present task begins earlier: it must recover the objects and relations needed to instantiate such a layout from one overhead image.

Executable visual reasoning offers a route between perception and structured assets. VISPROG and ViperGPT compose perception modules as inspectable programs \citep{gupta2023visprog,suris2023vipergpt}. LayoutGPT expresses spatial arrangements in code-like form, whereas 3D-GPT and Holodeck connect language-guided planning to 3D construction \citep{feng2023layoutgpt,sun2023threedgpt,yang2024holodeck}. Code-as-Room further generates executable indoor scenes from top-down images \citep{yang2026codeasroom}. Their settings center on general visual reasoning, language-conditioned generation, or indoor layouts. The framework instead addresses outdoor road networks and heterogeneous land cover, using one program for both a segmentation-compatible planar projection and a synchronized 3D scene.

%% file: chapters/03_method.tex
\section{Method}

\subsection{Task Contract and Semantic-Prior-Guided Parsing}

For a near-nadir RGB image $I$, the predicted artifact is a constrained program that executes into a typed city graph and a renderer-facing layout. The same named objects and geometry then yield an orthographic semantic projection and an oblique 3D view. Evaluation covers building, road area, sport, vegetation, and water. Within the program, \texttt{open\_space} is a generic renderable ground-surface object whose semantic type distinguishes vegetation, water, or sport; buildings and roads have dedicated constructors. Rendered objects use normalized geometry and typed relations, including road connectivity and spatial adjacency. Because the target is planimetric organization rather than metric 3D recovery from one view, height and appearance attributes are approximate.

An image model derives a canonical five-color semantic layout prior $M$ from $I$. Both inputs accompany every MLLM call: $M$ supplies aligned category and footprint evidence, while $I$ retains cues for function, relations, roof appearance, and approximate height. Parsing proceeds through three coordinated calls. The road pass recovers centerlines, widths, intersections, and connectivity. Given that road record, the topology pass extracts open-space regions and typed relations. The building pass then reads the accumulated topology to identify individual footprints and repeated groups. Coordinates remain normalized to the original image frame, and each later pass extends the shared object inventory rather than replacing it.

The prompts use $M$ as the principal planimetric constraint. After graph construction, connected components in the prior are used to reconcile the geometry of buildings, road areas, sports surfaces, and water. Building height and appearance fields still come from the RGB image, whereas vegetation and the remaining geometry are graph-derived. The prior therefore grounds MLLM interpretation without acting as an independent final segmentation head or as the reported prediction. Section~\ref{sec:prior-ablation} evaluates the same framework with and without this input.

\subsection{Typed Graph, Executable API, and Bounded Control}

Deterministic normalization converts the three records into stable identifiers, canonical coordinates, and typed relations for connectivity, containment, adjacency, service, and alignment. A serializer expresses this graph as Python code over a closed layout API. The main operations are \texttt{add\_road}, \texttt{add\_building}, \texttt{add\_open\_space}, and \texttt{relate}; schema-specific constructors additionally cover intersections and repeated building groups. Execution occurs in a timeout-limited subprocess that emits JSON, followed by checks for malformed geometry, duplicate identifiers, and invalid relation endpoints.

Recovery depends on the observed failure. Malformed MLLM JSON is salvaged or repaired; missing critical roads, buildings, open spaces, or relations can initiate a topology-focused retry before deterministic topology repair and validation. The controller consequently has a bounded operational role: scheduling specialized calls, retaining intermediate artifacts, executing the program, and sending failed checks to predefined recovery paths. The 100-scene quantitative evaluation uses this fixed pipeline.

\subsection{Shared Rendering and Evaluation}

One conversion turns the validated graph into a shared layout. Road centerlines are buffered by estimated widths, building polygons are extruded to approximate heights, and semantic open spaces become low surfaces. A deterministic Three.js renderer creates the oblique view; an orthographic projector rasterizes the identical object geometry with a fixed palette. For class $c$,
\begin{equation}
\operatorname{IoU}_{c}=\frac{|\hat{Y}_{c}\cap Y_{c}|}{|\hat{Y}_{c}\cup Y_{c}|},\qquad
\operatorname{mIoU}=\frac{1}{5}\sum_{c\in\mathcal{C}}\operatorname{IoU}_{c}.
\end{equation}
IoU is computed after class counts have been pooled across the evaluated scenes, preventing absent class--scene pairs from contributing artificial perfect scores. Since both views consume the same layout, an edit to a named object or relation is reflected when the 2D and 3D outputs are regenerated.

%% file: chapters/04_experiments.tex
\section{Experiments}

\subsection{Dataset and Evaluation Protocol}

The evaluation uses CityLayout-100, a fixed collection of 100 near-nadir $512\times512$ crops from IRSAMap V2 \citep{meng2025irsamap}. Difficulty is determined by the number of annotated land-cover instances: Easy scenes contain at most 17, Medium scenes 18--23, and Hard scenes at least 24. Scored classes are building, road area, sport, vegetation, and water; road-line annotations serve only as diagnostics. For each reported group, true positives, false positives, and false negatives are pooled over scenes before class IoU is calculated. mIoU averages the five pooled class IoUs, while GIoU, precision, recall, and F1 are obtained from counts pooled across classes. Thus an absent class--scene pair cannot introduce an artificial perfect score.

In the complete pipeline, \texttt{gpt-image-2} generates the semantic layout prior and \texttt{gpt-5.5} performs the decomposed MLLM parsing. The accumulated records subsequently pass through graph normalization, topology repair, constrained-code execution, Three.js construction, and orthographic evaluation.

\subsection{Overall and Class-wise Results}

\begin{table}[!t]
\centering
\caption{Complete-system aggregate results ($N=100$, \%).}
\label{tab:main}
\footnotesize
\setlength{\tabcolsep}{2.9pt}
\begin{tabular*}{\columnwidth}{@{\extracolsep{\fill}}lrrrrr@{}}
\toprule
Method & mIoU & GIoU & Prec. & Rec. & F1 \\
\midrule
Proposed framework & 41.1 & 48.3 & 65.8 & 64.6 & 65.2 \\
\bottomrule
\end{tabular*}
\end{table}

\begin{table}[!t]
\centering
\caption{Class-wise IoU of the complete system ($\%$).}
\label{tab:classwise}
\footnotesize
\setlength{\tabcolsep}{5.0pt}
\begin{tabular*}{\columnwidth}{@{\extracolsep{\fill}}lrrrrr@{}}
\toprule
Class & Build. & Road & Sport & Vege. & Water \\
\midrule
IoU & 44.8 & 14.0 & 34.2 & 53.1 & 59.5 \\
\bottomrule
\end{tabular*}
\end{table}

Once the program has executed and been projected, the framework attains 41.1\% mIoU and 48.3\% GIoU (Table~\ref{tab:main}). The executable representation therefore retains measurable agreement in the image plane. Water, vegetation, and buildings overlap more closely than road areas and sports surfaces (Table~\ref{tab:classwise}). Road-area IoU responds strongly to errors in buffered-network width and boundaries, even though the underlying roads remain named and connected rather than becoming anonymous pixels. These scores serve to assess the executable layout rather than to position a task-trained segmentation model.

\subsection{Complexity and Qualitative Analysis}

\begin{table}[!t]
\centering
\caption{Complete-system results by scene difficulty ($\%$).}
\label{tab:difficulty}
\footnotesize
\setlength{\tabcolsep}{10.0pt}
\begin{tabular*}{\columnwidth}{@{\extracolsep{\fill}}lrr@{}}
\toprule
Difficulty & mIoU & GIoU \\
\midrule
Easy & 42.3 & 48.5 \\
Medium & 36.3 & 50.9 \\
Hard & 42.7 & 45.6 \\
\bottomrule
\end{tabular*}
\end{table}

Instance-count difficulty does not produce a monotonic trend in Table~\ref{tab:difficulty}. Medium scenes yield the highest GIoU but the lowest mIoU, consistent with good overlap on dominant areas and less even agreement across classes. Hard scenes still reach 42.7\% mIoU, suggesting that boundary ambiguity and rare-class geometry influence the result independently of the number of instances.

\begin{figure*}[!t]
    \centering
    \includegraphics[width=0.95\textwidth]{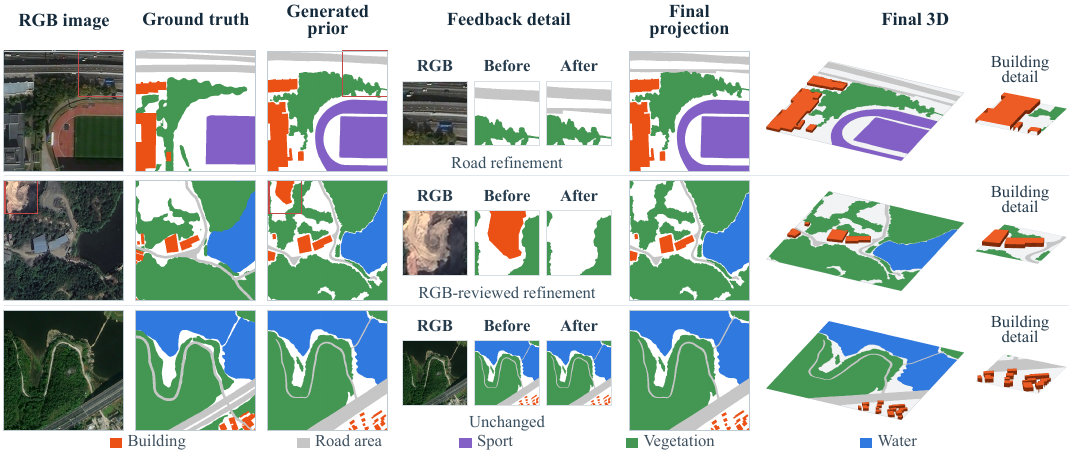}
    \caption{Qualitative feedback examples with RGB/before/after details and building close-ups. Case 2 uses the RGB-reviewed mask. Final projections and 3D layouts share each case's final mask; building heights are illustrative.}
    \label{fig:qualitative}
\end{figure*}

Fig.~\ref{fig:qualitative} shows RGB-guided local refinement and the resulting planar and 3D layouts. Case 2 incorporates RGB-reviewed building corrections. The final mask of each case generates both views. Tables~\ref{tab:main}--\ref{tab:prior-ablation} report the 100-scene pipeline evaluation.

\subsection{Ablation of the Semantic Layout Prior}
\label{sec:prior-ablation}

\begin{table}[!t]
\centering
\caption{Effect of the generated semantic layout prior ($N=100$, \%).}
\label{tab:prior-ablation}
\footnotesize
\setlength{\tabcolsep}{2.8pt}
\begin{tabular*}{\columnwidth}{@{\extracolsep{\fill}}lrrrrr@{}}
\toprule
Variant & mIoU & GIoU & Prec. & Rec. & F1 \\
\midrule
Without semantic layout prior & 36.3 & 42.5 & 60.7 & 58.7 & 59.7 \\
With semantic layout prior & \textbf{41.1} & \textbf{48.3} & \textbf{65.8} & \textbf{64.6} & \textbf{65.2} \\
\bottomrule
\end{tabular*}
\end{table}

Table~\ref{tab:prior-ablation} isolates the generated semantic layout prior by evaluating the same pipeline with and without that input. In its absence, the MLLM parses the RGB image alone; when present, aligned class and footprint evidence grounds each parsing pass. The difference is 4.8 points in mIoU, 5.8 points in GIoU, and 5.5 points in F1. Per-class IoU changes by $+17.6$ for building, $+4.0$ for road area, $-8.3$ for sport, $+2.7$ for vegetation, and $+8.4$ for water. The aggregate gain supports semantic grounding for MLLM layout reasoning, while the sport result shows that an inaccurate prior may also propagate into executable geometry.

%% file: chapters/05_conclusion.tex
\section{Conclusion}

The framework recovers a measurable urban layout by generating constrained code with an MLLM. Localization is retained in named objects, the orthographic projection permits segmentation-style evaluation, and the executable prediction keeps typed relations and editable construction operations together with a synchronized 3D output. Over 100 scenes, the generated semantic layout prior improves the alignment of executed layouts by grounding decomposed MLLM reasoning and selected footprints. In this form, remote-sensing detection and segmentation can feed an executable-layout representation whose code remains measurable, inspectable, editable, and renderable. Subject to human review, the structured objects and geometry could be translated into city-data formats and adapted for OSM2World-like map-to-3D asset maintenance.